# Hybrid lemmatization in HuSpaCy

Péter Berkecz, György Orosz, Zsolt Szántó,
Gergő Szabó, Richárd Farkas

Institute of Informatics, University of Szeged
2. Árpád tér, Szeged, Hungary
{berkecz,szantozs,gszabo,rfarkas}@inf.u-szeged.hu
gyorgy@orosz.link

**Abstract.** Lemmatization is still not a trivial task for morphologically rich languages. Previous studies showed that hybrid architectures usually work better for these languages and can yield great results. This paper presents a hybrid lemmatizer utilizing both a neural model, dictionaries and hand-crafted rules. We introduce a hybrid architecture along with empirical results on a widely used Hungarian dataset. The presented methods are published as three HuSpaCy models.

## 1 Introduction

Lemmatization is the process where the lemma - the dictionary form of a word - needs to be computed from an inflected word form. This can happen in various ways. One approach is that we are deriving the possible lemmata of a token using a morphological analyzer which relies on hand-crafted morphological rules then a statistical model is applied to disambiguate using the context. On the other hand, end-to-end statistical lemmatizers are using a training corpus exclusively to learn the transformation rules and the context disambiguation in a single model. Hence, they can track the language evolution and fit domain variations if a training dataset is available.

Several end-to-end statistical lemmatizers have been proposed (Müller et al., 2015; Straka, 2018) however, based on our experiments on the Hungarian subset of Universal Dependencies (Nivre et al., 2017), they were not accurate enough on their own. Studies (Orosz and Novák, 2013; Boudchiche and Mazroui, 2019) showed, that for morphologically rich languages hybrid architectures tend to perform better.

In this paper, we propose the following three-step hybrid architecture:

1. a dictionary-based first step for unambiguous cases, where we just replace the token with the lemma if the traits are the same,
2. end-to-end statistical lemmatizer for the highest accuracy in ambiguous cases
3. hand-made rules for special cases where the statistical approach did not see enough examples.

For end-to-end lemmatization, we investigated two algorithms that learn rules from training data, the Lemmy (Kristiansen, 2019) and the Edit-tree Lemmatizer (de Kok, 2021). Both of them use neural networks to find the best set of rules for a given word in a given context. We examined various word embedding and transformer-based networks to find the best-performing end-to-end lemmatizer for Hungarian.

Our tool is built on spaCy's foundation and is both versatile and easy to use in a multi-language environment. It is conveniently released under a Creative Commons BY-SA 4.0 license, making it an attractive option for organizations looking to incorporate natural language processing capabilities into their systems without any licensing issues.

Our hybrid lemmatizer is available in the open-source HuSpaCy toolkit, in the `medium`, `large`, and `transformer` models since version 3.4.

## 2 Related Work

### 2.1 Lemmatizing in spaCy

spaCy (Honnibal, 2015) is an industrial-strength natural language processing pipeline, which already supports lemmatization for nineteen languages. HuSpaCy (Orosz et al., 2022) is a Python package supporting Hungarian models for spaCy. Three different models are available for the tool, with two different embedding types. The `medium` and `large` model uses static word vectors, while the `transformer` model relies on `huBERT` (Nemeskey, 2020a), a fine-tuned BERT (Devlin et al., 2018) model for Hungarian. Up until now, HuSpaCy utilized a simple rule-learning algorithm it used a lemmatizer called Lemmy.

There are three different built-in lemmatizers available in spaCy. The first is the ruled-based one. It uses hand-crafted rules and part-of-speech tags to make a lemma from a token. Four official spaCy models use this, English, French, Spanish, and Macedonian.

The second type is a dictionary-based lemmatizer. It stores the tokens with their PoS tags, and it loads a lemma for a token during prediction. Only one official spaCy model uses this, the Catalan one.

The third and newest one is the Edit-tree Lemmatizer, which is an end-to-end neural lemmatizer. Fourteen official spaCy models are using this, for example, Greek, Polish, Danish, German, and Finnish.

### 2.2 Hungarian Lemmatization Approaches

In Hungarian, there are several approaches to lemmatizing texts, for example, `magyarlanc` (Zsibrita et al., 2013) and `emtsv` (Simon et al., 2020; Indig et al., 2019a,b). Both of them use symbolic methods to make lemma candidates and disambiguate between them based on the context of the word in another step.

The morphological parser in `magyarlanc` generates lemma candidates based on KR (Trón et al., 2006) grammatical rules and then uses the PurePOS part-of-speech tagger (Orosz and Novák, 2013) to select the correct one according to the sentence context.

`emtsv` is the result of a collective effort to integrate multiple existing NLP tasks into one application. For lemmatization, it uses the emMorph (Novák et al., 2016; Novák, 2014) morphological analyzer to create a lemma, part-of-speech tag, and morphological description candidates, and the emTag (PurePOS) tagger to disambiguate between them. During prediction, it uses Hungarian rules (Novák, 2015) for HFST (Lindén et al., 2009), then from the pool of lemma candidates, it selects the most probable one using the PurePOS tagger. We must note that emMorph has a very restrictive license, thus it cannot be used freely in a commercial application.

The lemmatizer of Stanza (Qi et al., 2020) combines a dictionary-based lemmatizer with a neural Seq2Seq lemmatizer. The dictionary is being used as a cache to predict lemmata quickly because the Seq2Seq lemmatizer is typically slower, this is also compiled from the training dataset.

UDPipe (Straka, 2018) handles lemmatization as a sub-task of tagging. The system generates a rule from tokens and their lemmata based on the longest common subsequence of two words and then selects them as a classification task.

## 3 Methods

Various studies (Orosz et al., 2022; Jongejan and Dalianis, 2009) have already shown how the different types of lemmatizers perform individually. In some cases of agglutinating languages, however, the best solution was usually achieved with a combination of these (Orosz, 2015). In the following section, we present the methods we have selected for our architecture.

### 3.1 Lemmy

Lemmy is an end-to-end statistical lemmatizer. It is an open-source Python implementation of the CST algorithm (Jongejan and Dalianis, 2009). This method automatically learns rules from the training data by detecting prefixes and suffixes to be transformed, so that the resulting token can be used to generate the lemma. First, it tries to lemmatize based on the existing rules, if this fails, it creates a new rule using their longest common subsequence and their differences. It creates a rule based on the token and its lemma's longest common subsequence, and stores this and the token's PoS-tag with the rule occurrence count. It had a few modifications over the original project to improve its accuracy. During prediction, it uses this frequency information and the predicted PoS-tag to select a rule, then applies it to the token to compute the lemma.

### 3.2 Edit-tree Lemmatizer

The Edit-tree Lemmatizer is another end-to-end statistical lemmatizer that exploits neural language models. It is a spaCy implementation of the Lemming lemmatizer (Müller et al., 2015), which learns rules (called edit-trees) from the training data to transform a word into its lemma.

For example, this system builds an edit-tree for the word *leghosszabb* (longest) this way:

1. Search for the longest common subsequence (LCS) between the token and its lemma: *leghosszabb* - *hosszú* (long)
2. Based on this, it splits into 3 parts, prefix - LCS - suffix
3. Finds the necessary transformations for the prefix and suffix to transform the inflected form into the lemma
   (a) In the prefix, we change *leg-* into an empty string ($\epsilon$)
   (b) In the suffix, we change *-abb* into *-ú*

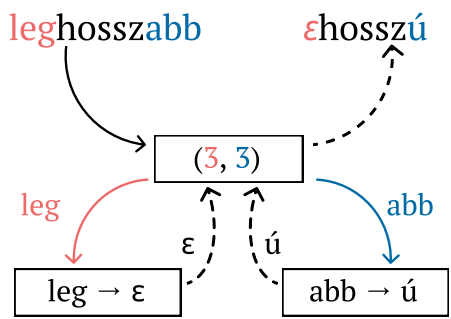

Fig. 1: Edit-tree for the word *leghosszabb*

When predicting, a neural network is used as a classifier to select the right edit-tree. In spaCy we can use multi-task learning and this can be inserted as a task, thus being able to access hidden representations in the underlying neural network. This way, for example, the tagger and the lemmatizer can work together in an even closer relationship, not just on the few tags that the tagger predicts.

### 3.3 Hungarian Language Models Used

The architecture of HuSpaCy (Orosz et al., 2022) consisted of a Word2Vec (Mikolov et al., 2013) static word embedding and CNN layers. The disadvantage of Word2Vec is that it only knows the words it saw during training and does not use the context information of the tokens.

To overcome these drawbacks, we experimented with word embeddings containing subword information and context-aware transformer networks.

Static word embeddings containing subword information (e.g. `fastText` (Bojanowski et al., 2017)) learn not just a single word during training, but several small parts of a word. This way, because a token is a sum of its character *n-gram* vectors, it performs better for words outside the vocabulary (OOV), be it a truly unknown word or a typo. It has the advantage of having a small storage footprint due to the smaller vocabulary size and because of that, it is quite fast when evaluated on the CPU. In our experiments, we have utilized `floret` (Boyd, 2021) which is the spaCy fork of the `fastText`. We trained two models on the Hungarian WebCorpus 2.0 (Nemeskey, 2020b), a 100-dimensional and a 300-dimensional one. The advantage of them is that they include character n-grams, and due to the hashing trick of `floret`, their storage size is much smaller than the original word embeddings.

On the other hand, transformer models (e.g. `BERT`) are able to take the full sentence context of the word into consideration when predicting, and thus can give a more accurate representation and achieve a high level of accuracy in HuSpaCy (Szabó et al., 2023). Their disadvantages are their large storage size and relatively slow CPU evaluation, but with GPU it can be fast.

### 3.4 Lookup Lemmatizer

Dictionary-based lemmatizers only perform a simple replacement operation. They "learn" to map tokens to lemmata from the training dataset by memorizing a word form with their morphosyntactic information associated with lemmata. If more than one lemma is associated with the same word with the same morphosyntactic labels, the most frequent one is learned. In prediction time, they simply return the matching lemmata if any. The disadvantage is that it can only lemmatize tokens that it has already seen.

### 3.5 Postprocessing Rules

There may be cases where the lemmatizer makes mistakes and these cases can be easily corrected by rules. These special hand-crafted rules are most often used when something has to be cut off the end of a token to get the correct lemma. You could actually call it the little brother of stemming. Beyond suffix cutting, these rules are also useful in certain cases for agglutinating languages.

We noticed that in a few rare cases where there are exclamation and question marks in a token they were left in a lemma, thus we've built a rule to remove them. Also, there were some numbers and dates that needed to be fixed. For example, for the token *4-6-os* the lemmatizers usually made *4-6-* as a lemma, but *4-6* would be the correct one. Because the predicted lemmata had unrecognizable patterns, this rule used the original token to remove any suffixes after a number or a date.

## 4 Results

### 4.1 Experimental Setup

To ensure comparability, we have used the most commonly available tools for training and testing. The evaluations were carried out on the test set of the Hungarian Universal Dependencies (UD) (Nivre et al., 2017) corpus using the CoNLL 2018 Shared Task (Zeman et al., 2018) evaluator script[1].

For training, we used parts of the Szeged Korpusz (Csendes et al., 2004) that are not included in any part of the UD corpus. As for the comparison, we chose the most popular open-source tools and used them as they were published for reproducibility.

### 4.2 Experiments

Lemmy was used in HuSpaCy, but it only relied on the part-of-speech information and did not have any other information about the token. The Edit-tree Lemmatizer was released in spaCy 3.3. Studies (de Kok, 2021) showed that it brings significant improvements for various languages, hence we investigated its usage for Hungarian (cf. Table 1).

| Lemmy | Edit-tree Lemmatizer |
|---|---|
| 95.53% | 95.90% |

**Table 1.** Results of the previous Lemmy lemmatizer and the Edit-tree Lemmatizer with the default configurations. We consider this as the baseline.

As we have shown in subsection 3.3, the underlying neural language model can be simply replaced by another spaCy-compatible module. The Word2Vec word embeddings previously shipped with HuSpacy models were replaced by a `floret` (Boyd, 2021) architecture. This not only improves our results but also reduces the storage requirements of the whole model.

We experimented with pretrained transformer language models as well. Based on our experiments (cf. Table 2), huBERT (Nemeskey, 2020a) gave a good accuracy-to-speed ratio, but with an `XLM-Roberta-Large` (Conneau et al., 2019) model we were able to achieve even higher accuracy at the expense of throughput.

| `floret`(100d) | `floret`(300d) | huBERT | XLM-Roberta |
|---|---|---|---|
| 96.56% | 96.76% | 98.53% | 98.89% |

**Table 2.** Results of the Edit-tree Lemmatizer with different language models.

---
[1] https://universaldependencies.org/conll18/conll18_ud_eval.py

The Edit-tree Lemmatizer has an important hyperparameter, which controls the number of edit-trees to try before giving up and choosing another one of the token's properties as a lemma. It defaults to 1, but increasing this value improved the accuracy. By adjusting this, the model can access more edit-tree candidates during prediction because the possible edit-tree is chosen from all of them. Increasing it to 3 improves the result (cf. Table 3) even though the pipeline is slower, but it's a worthy compromise. A higher `top_k` value makes the algorithm a bit slower (cf. Table 4) and has almost no effect on the results. Because of this, we've choosen 3 as our `top_k` value for the next experiments.

|  | floret(100d) | floret(300d) | huBERT | XLM-Roberta |
|---|---|---|---|---|
| `top_k` = 1 | 96.56% | 96.76% | 98.53% | 98.89% |
| `top_k` = 3 | 96.87% | 97.01% | 98.63% | 98.93% |
| `top_k` = 8 | 96.84% | 97.09% | 98.61% | 98.83% |

**Table 3.** Lemma accuracies of the Edit-tree Lemmatizer with increased `top_k` parameters.

|  | floret(100d) | floret(300d) | huBERT | XLM-Roberta |
|---|---|---|---|---|
| `top_k` = 1 | 7739 | 7255 | 3349 | 2429 |
| `top_k` = 3 | 6740 | 6873 | 3166 | 2358 |
| `top_k` = 8 | 6655 | 6571 | 3146 | 2347 |

**Table 4.** Throughput (token/s) of the Edit-tree Lemmatizer with increased `top_k` parameters.

In Lemmy, it was already noticeable that the casing of some words was wrong, this problem was also apparent here. So the same rule had to be built into the system (cf. Table 5). If the word is a proper noun or the first token of a sentence, it can start with a capital letter, in all other cases, the word is lowercase. We built this in so that the algorithm can be aware of these casings during training so that it can apply them later during prediction.

|  | floret(100d) | floret(300d) | huBERT | XLM-Roberta |
|---|---|---|---|---|
| + Casing | 96.29% | 97.23% | 98.63% | 98.85% |

**Table 5.** Results of the Edit-tree Lemmatizer taking casing into account.

**Hybrid Lemmatizer Results** Since the Edit-tree Lemmatizer doesn't give any prediction sometimes, we tested how a dictionary-based lemmatizer performs as a supplement. This became the second part of our architecture, but it became a pipeline step before the Edit-tree Lemmatizer in the whole spaCy pipeline.

We had two experiments with this setup, the first was about pairing the training data to the lemmata based on the token and PoS tag (cf. Table 6). However, this led to many incorrect predictions. Finally, to solve this, we paired the training data to the lemmata to be trained based on the token, the PoS-tag, and the morphological tags. During training and prediction, the number-like tokens are masked, resulting in fewer pairs, but this will match more tokens containing numbers. It can generalize better with this because for example the *1000-ben* will be *0000-ben* and it could be applied to *3000-ben* which may not have been included in the training set.

In a couple of cases, even combined – the dictionary-based lemmatizer with the Edit-tree Lemmatizer – they still made some errors, and we used post-processing rules to try to correct the lemma in the ways that we mention in subsection 3.5. Upon implementation of these rules, the accuracies demonstrated a slight decrease, however, the overall quality of the predicted lemmata showed noticeable improvement.

|          | floret(100d) | floret(300d) | huBERT | XLM-Roberta |
|----------|--------------|--------------|--------|-------------|
| + Lookup | 97.19%       | 97.48%       | 98.68% | 98.94%      |
| + Rules  | 97.17%       | 97.46%       | 98.68% | 98.94%      |

**Table 6.** Results of the hybrid architecture. The rows are an addition to the existing system. It can be observed that the transformer-based models were not really helped by the hybrid additions, but the floret-based models were. This is due to the fact that the larger models are more adept at selecting the edit-trees.

### 4.3  Final Results

In order to compare our architecture's results with already existing systems, we chose three popular systems that have built-in lemmatizers. emtsv is a Hungarian pipeline of state-of-the-art NLP components, UDPipe is commonly used as a baseline in CoNLL competitions, and Stanza has relatively high scores on UD test sets. It is important to point out that all of them were used off-the-shelf, i.e. not re-trained or fine-tuned.

The evaluation was performed on the Hungarian Universal Dependencies test set. The UDPipe and Stanza authors have made the results of their systems available, but the emtsv authors have not, so we used one of the default configurations to get UD-compatible output and then evaluate it. It is important to point out that comparing emtsv's lemmatizer is probably not fair out-of-the-box, because the tool was trained on a different train-test split, which may conflict with ours. (There is a high probability that their training set contains sentences in our test set.)

In Table 7 one can see that all systems are quite accurate except UDPipe. Our hybrid lemmatizer consistently outperforms other systems, demonstrating a marked improvement in performance.

|  | Accuracy |
|---|---|
| Word2Vec *baseline* | 95.90% |
| HuSpaCy + `floret` (100d) | 97.17% |
| HuSpaCy + `floret` (300d) | 97.46% |
| HuSpaCy + `huBERT` | 98.68% |
| HuSpaCy + `XLM-Roberta` | 98.94% |
| `emtsv` | 96.16% |
| UDPipe | 88.50% |
| Stanza | 94.19% |

**Table 7.** Lemmatization accuracies of the different NLP pipelines.

### 4.4 Resource Usage

It is also important to mention the resource usage of our models. These models contain a complete language pipeline, i.e. a single model of a system. In the case of HuSpaCy, these models include tagger, morphologizer, lemmatizer, parser, and named entity recognition.

|  | Storage Size | Throughput (tokens/sec) CPU | Throughput (tokens/sec) GPU | Memory Usage (GiB) |
|---|---|---|---|---|
| Word2Vec *baseline* | 1.6 GB | 854 | 7059 | 4.61 |
| HuSpaCy + `floret` (100d) | 125 MB | 1862 | 5903 | 2.35 |
| HuSpaCy + `floret` (300d) | 455 MB | 1186 | 6074 | 3.33 |
| HuSpaCy + `huBERT` | 1.8 GB | 242 | 3055 | 4.84 |
| HuSpaCy + `XLM-Roberta` | 9.1 GB | 78 | 2186 | 17.87 |
| `emtsv` | 3.6 GB | 116 | - | 3.91 |
| UDPipe | 5 MB | 3175 | - | 1.38 |
| Stanza | 246 MB | 30 | 395 | 5.31 |

**Table 8.** Resource usages of the different models. The storage sizes contain a full natural language processing pipeline, not just the lemmatizer. During measurement, we've tried to run every benchmark three times, and take the best result with each tool. For each tool, we've tried to use only the parts that are needed to the lemmatizer, in the case of our models this means that we've run them without named entity recognition. We measured the throughput using an AMD EPYC 7F72 CPU and NVIDIA A100 (40GB) GPU.

Table 8 shows that the results do not really correlate with the storage requirements of a model. This is mostly due to the language model, as can be seen in the case of the Word2Vec model, which is 1.6GB but is outperformed by a 125MB model in accuracy. However, the two models use completely differ-ent word embeddings, while Word2Vec uses a continuous bag-of-words model, `floret` uses CBOW with subword information but also hashes words so it is a bit slower.

For both `XLM-Roberta`-based and `huBERT`-based, we can see that the 5x storage requirement is almost unnecessary, with the result being only a few tenths of a percentage point better.

We must note that at `emtsv` it is the size of a Docker image which is a recommended way to use the tool, thus it may contain parts that are not part of the original `emtsv` pipeline.

## 5 Conclusions

We presented a hybrid architecture, which consists of three parts utilizing the advantages of neural language model-based statistical lemmatizers and dictionaries along with hand-crafted rules. We have shown the application of hand-crafted rules and morphological information-based dictionaries can improve end-to-end statistical lemmatizers. We examined different language models and we found the `XLM-Roberta-large` model achieves the best results, but the five times smaller `huBERT` reached comparable results too. We performed multiple experiments to extensively compare our tool with other systems. As the results show, our lemmatizer outperforms other models and the storage requirement is smaller with the smaller `floret` model than some of them.

Altogether, this paper has presented a new hybrid lemmatizer architecture that is freely available in three models of HuSpaCy. We made it open-source, under a permissive license.

## Acknowledgements

The authors would like to thank Dávid Nemeskey and Dániel Lévai for their help in benchmarking `emBERT` and Stanza. HuSpaCy research and development is supported by the European Union project RRF-2.3.1-21-2022-00004 within the framework of the Artificial Intelligence National Laboratory.

XIX. Magyar Számítógépes Nyelvészeti Konferencia		Szeged, 2023. január 26–27.